\documentclass{article}
\usepackage{graphicx}
\usepackage{booktabs}
\usepackage{multirow}
\usepackage{caption}
\usepackage{adjustbox}
\usepackage{amsmath}
\usepackage{floatflt}
\usepackage{wrapfig}
\usepackage{hyperref}
\usepackage{url}
\usepackage{amsmath}
\usepackage{amssymb}
\usepackage{booktabs}    
\usepackage{graphicx}    
\usepackage{hyperref}    
\usepackage{enumitem}    
\usepackage{xcolor}

\usepackage{graphicx}
\usepackage[svgnames]{xcolor}
\usepackage{tikz} 
\usepackage[most]{tcolorbox}
\tcbuselibrary{skins}

\newcommand{\helv}{\fontfamily{phv}\selectfont}

\newcommand{\resultfont}{\helv\bfseries\small}

\newtcolorbox{innerbox}{
    colback=gray!5, 
    colframe=gray!10, 
    boxrule=0.5pt, 
    arc=2pt,
    left=3pt, right=3pt, top=3pt, bottom=3pt,
    enhanced,
    width=\linewidth
}

\tcbset{
    myboxstyle/.style={
        enhanced,
        colback=white,
        colframe=gray!40,
        coltitle=black,
        fonttitle=\helv\bfseries\small, 
        boxrule=1pt,
        arc=3pt,
        left=5pt, right=5pt, top=5pt, bottom=5pt,
        equal height group=A, 
    }
}

\definecolor{startblue}{rgb}{0.53, 0.81, 0.98} 
\definecolor{endblue}{rgb}{0.0, 0.0, 0.549}   

\newcommand{\nop}[1]{}

\PassOptionsToPackage{numbers, compress}{natbib}
\usepackage[preprint]{neurips_2024}

\title{Learning When Not to Attend Globally}
\author{Xuan Luo, Kailai Zhang, Xifeng Yan \\
Department of Computer Science, UC Santa Barbara \\
\{xuan\_luo, kailai, xyan\}@cs.ucsb.edu}

\begin{document}

\maketitle

\begin{abstract}
When reading books, humans focus primarily on the current page, flipping back to recap prior context only when necessary. Similarly, we demonstrate that Large Language Models (LLMs) can learn to dynamically determine when to attend to global context. We propose All-or-Here Attention (AHA), which utilizes a binary router per attention head to dynamically toggle between full attention and local sliding window attention for each token. Our results indicate that with a window size of 256 tokens, up to 93\% of the original full attention operations can be replaced by sliding window attention without performance loss. Furthermore, by evaluating AHA across various window sizes, we identify a long-tail distribution in context dependency, where the necessity for full attention decays rapidly as the local window expands. By decoupling local processing from global access, AHA reveals that full attention is largely redundant, and that efficient inference requires only on-demand access to the global context.
\end{abstract}

\section{Introduction}

The self-attention mechanism~\cite{attention} serves as the cornerstone of the Transformer architecture~\cite{bert,gpt,transformer}, driving the success of modern Large Language Models~\cite{instructgpt,llama,qwen3}. However, its quadratic complexity with respect to sequence length poses significant computational challenges. To address this, numerous studies have proposed efficiency techniques, including linear attention~\cite{linearattn,mamba,gateddeltanet}, sparse attention~\cite{sparsetransformer,longformer,bigbird}, and hierarchical attention~\cite{hierarchical,deepseeknsa}. While these methods often focus on approximating full attention or compressing state representations, we suggest that the intrinsic redundancy of the attention mechanism enables a far simpler solution.

In this paper, we propose a straightforward paradigm: enabling the Transformer to \textit{learn when not to attend globally}. Our design is motivated by the observation that tokens exhibit heterogeneous context requirements. Intuitively, maintaining syntactic continuity and local coherence should primarily require immediate context, whereas global retrieval is necessary only for a sparse subset of tokens resolving long-range dependencies~\cite{dependency1,dependency2,dependency3,longformer}. Based on this perspective, we introduce \textbf{All-or-Here Attention (AHA)}. By employing a hard gating mechanism, AHA dynamically toggles between full attention (``All'') and local sliding window attention (``Here'') for each attention head. Specifically, we integrate lightweight routers at each Transformer layer to generate scalar importance scores. Based on these scores and a pre-defined threshold, each attention head determines whether to execute full or sliding window attention. This represents a fundamental departure from methods such as gated attention~\cite{gateddeltanet, gatedattention} and hierachical attention~\cite{deepseeknsa, hierarchical,deepseekv32}, which adopt soft gating to continuously re-weight and fuse the features. In contrast, our approach implements conditional computation, treating context selection as a strict binary decision to explicitly decouple local processing from global retrieval.

Our experiments demonstrate that the vast majority of full attention operations in standard LLMs are not needed. Using OLMo-2~\cite{olmo2} as our base model, we show that with a window size of 256, over 93\% of full attention operations can be substituted with local sliding window attention while maintaining full performance on standard benchmarks. We further analyze the relationship between window size and attention sparsity, revealing that context dependency follows a long-tail distribution. As the local window expands, the necessity for global access decays rapidly, dropping from 53\% at a window size of 16 to under 7\% at a window size of 256. These findings validate our hypothesis that full attention is largely unnecessary. By explicitly decoupling local processing from global retrieval, AHA offers a minimalist yet powerful paradigm for efficient Large Language Models.

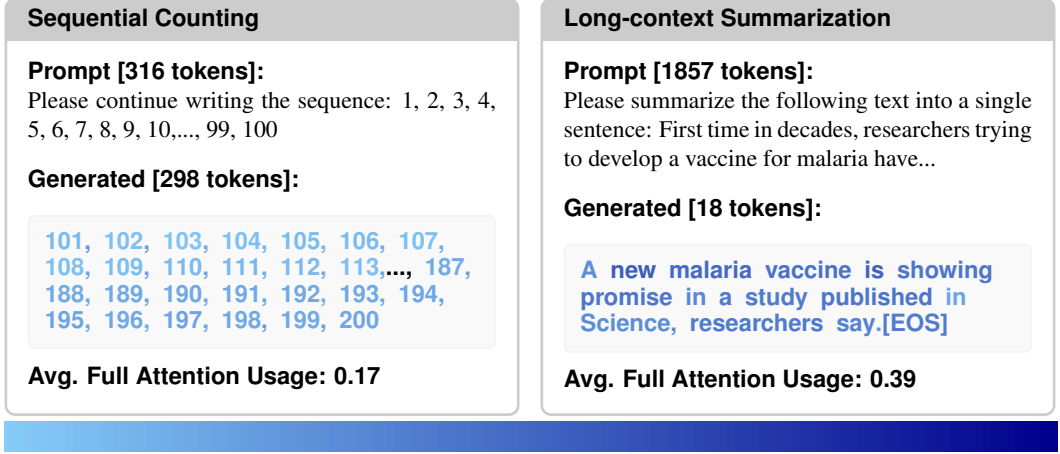
\begin{figure}[t]
    \centering
    
    \noindent
    \begin{minipage}[t]{0.49\textwidth} 
        \begin{tcolorbox}[myboxstyle, title=Sequential Counting]
            {\helv\bfseries\small Prompt [316 tokens]:} \par 
            {\small Please continue writing the sequence:
1, 2, 3, 4, 5, 6, 7, 8, 9, 10,..., 99, 100}
            \par\vspace{0.8em}
            
            {\helv\bfseries\small Generated [298 tokens]:} \par
            \vspace{0.2em}
            \begin{innerbox}
                \resultfont
{\fontfamily{phv}\selectfont\bfseries\small
\textcolor[rgb]{0.477,0.729,0.937}{101}\textcolor[rgb]{0.382,0.583,0.859}{,}\hspace{1ex}\textcolor[rgb]{0.493,0.753,0.950}{102}\textcolor[rgb]{0.408,0.624,0.881}{,}\hspace{1ex}\textcolor[rgb]{0.504,0.770,0.958}{103}\textcolor[rgb]{0.429,0.656,0.898}{,}\hspace{1ex}\textcolor[rgb]{0.493,0.753,0.950}{104}\textcolor[rgb]{0.461,0.705,0.924}{,}\hspace{1ex}\textcolor[rgb]{0.482,0.737,0.941}{105}\textcolor[rgb]{0.461,0.705,0.924}{,}\hspace{1ex}\textcolor[rgb]{0.482,0.737,0.941}{106}\textcolor[rgb]{0.435,0.664,0.902}{,}\hspace{1ex}\textcolor[rgb]{0.493,0.753,0.950}{107}\textcolor[rgb]{0.440,0.672,0.907}{,}\hspace{1ex}\textcolor[rgb]{0.493,0.753,0.950}{108}\textcolor[rgb]{0.424,0.648,0.894}{,}\hspace{1ex}\textcolor[rgb]{0.488,0.745,0.946}{109}\textcolor[rgb]{0.419,0.640,0.889}{,}\hspace{1ex}\textcolor[rgb]{0.477,0.729,0.937}{110}\textcolor[rgb]{0.419,0.640,0.889}{,}\hspace{1ex}\textcolor[rgb]{0.472,0.721,0.933}{111}\textcolor[rgb]{0.440,0.672,0.907}{,}\hspace{1ex}\textcolor[rgb]{0.482,0.737,0.941}{112}\textcolor[rgb]{0.435,0.664,0.902}{,}\hspace{1ex}\textcolor[rgb]{0.493,0.753,0.950}{113}\textcolor[rgb]{0.424,0.648,0.894}{,}...{,}\hspace{1ex}\textcolor[rgb]{0.440,0.672,0.907}{187}\textcolor[rgb]{0.413,0.632,0.885}{,}\hspace{1ex}\textcolor[rgb]{0.440,0.672,0.907}{188}\textcolor[rgb]{0.429,0.656,0.898}{,}\hspace{1ex}\textcolor[rgb]{0.435,0.664,0.902}{189}\textcolor[rgb]{0.419,0.640,0.889}{,}\hspace{1ex}\textcolor[rgb]{0.435,0.664,0.902}{190}\textcolor[rgb]{0.429,0.656,0.898}{,}\hspace{1ex}\textcolor[rgb]{0.435,0.664,0.902}{191}\textcolor[rgb]{0.424,0.648,0.894}{,}\hspace{1ex}\textcolor[rgb]{0.435,0.664,0.902}{192}\textcolor[rgb]{0.451,0.689,0.915}{,}\hspace{1ex}\textcolor[rgb]{0.435,0.664,0.902}{193}\textcolor[rgb]{0.424,0.648,0.894}{,}\hspace{1ex}\textcolor[rgb]{0.429,0.656,0.898}{194}\textcolor[rgb]{0.403,0.616,0.877}{,}\hspace{1ex}\textcolor[rgb]{0.424,0.648,0.894}{195}\textcolor[rgb]{0.429,0.656,0.898}{,}\hspace{1ex}\textcolor[rgb]{0.435,0.664,0.902}{196}\textcolor[rgb]{0.445,0.680,0.911}{,}\hspace{1ex}\textcolor[rgb]{0.424,0.648,0.894}{197}\textcolor[rgb]{0.424,0.648,0.894}{,}\hspace{1ex}\textcolor[rgb]{0.435,0.664,0.902}{198}\textcolor[rgb]{0.398,0.608,0.872}{,}\hspace{1ex}\textcolor[rgb]{0.435,0.664,0.902}{199}\textcolor[rgb]{0.408,0.624,0.881}{,}\hspace{1ex}\textcolor[rgb]{0.429,0.656,0.898}{200}
}
            \end{innerbox}
            
            \par\vfill 
            {\helv\footnotesize\bfseries Avg. Full Attention Usage: 0.17}
        \end{tcolorbox}
    \end{minipage}%
    \hfill
    \begin{minipage}[t]{0.49\textwidth}
        \begin{tcolorbox}[myboxstyle, title=Long-context Summarization]
            {\helv\bfseries\small Prompt [1857 tokens]:} \par
            {\small Please summarize the following text into a single sentence: First time in decades, researchers trying to develop a vaccine for malaria have...}
            \par\vspace{0.8em}
            
            {\helv\bfseries\small Generated [18 tokens]:} \par
            \vspace{0.2em}
            \begin{innerbox}
                \resultfont
                \textcolor[rgb]{0.366,0.559,0.846}{A}\hspace{1ex}\textcolor[rgb]{0.238,0.364,0.743}{new}\hspace{1ex}\textcolor[rgb]{0.270,0.413,0.769}{malaria}\hspace{1ex}\textcolor[rgb]{0.291,0.446,0.786}{vaccine}\hspace{1ex}\textcolor[rgb]{0.254,0.389,0.756}{is}\hspace{1ex}\textcolor[rgb]{0.302,0.462,0.795}{showing}\hspace{1ex}\textcolor[rgb]{0.302,0.462,0.795}{promise}\hspace{1ex}\textcolor[rgb]{0.350,0.535,0.833}{in}\hspace{1ex}\textcolor[rgb]{0.313,0.478,0.803}{a}\hspace{1ex}\textcolor[rgb]{0.318,0.486,0.808}{study}\hspace{1ex}\textcolor[rgb]{0.302,0.462,0.795}{published}\hspace{1ex}\textcolor[rgb]{0.424,0.648,0.894}{in}\hspace{1ex}\textcolor[rgb]{0.360,0.551,0.842}{Science}\textcolor[rgb]{0.350,0.535,0.833}{,}\hspace{1ex}\textcolor[rgb]{0.313,0.478,0.803}{researchers}\hspace{1ex}\textcolor[rgb]{0.334,0.510,0.821}{say}\textcolor[rgb]{0.376,0.575,0.855}{.}\textcolor[rgb]{0.318,0.486,0.808}{[EOS]}
            \end{innerbox}
            
            \par\vfill 
            {\helv\footnotesize\bfseries Avg. Full Attention Usage: 0.39}
        \end{tcolorbox}
    \end{minipage}

    \par\vspace{0.1em} 
    \centering
    \begin{tikzpicture}
        \shade[left color=startblue, right color=endblue] (-1, -0.20) rectangle (13, 0.20);
    \end{tikzpicture}

    \caption{Visualization of full average attention usage across tasks (local attention window size,  $w=128$). The color gradient indicates the ratio of activating full attention, where dark blue represents 1.0 (Full Attention) and light blue represents 0.0 (Local Attention). This highlights that reliance on global context varies by task nature and token characteristics.}
    \label{fig:fig1}
\end{figure}

\section{Method}

In this section, we detail the proposed All-or-Here Attention (AHA), which introduces a conditional computation mechanism into the attention layer. The core intuition is to assign a binary gate to each attention head, dynamically routing tokens to either a full ('all') or a local sliding window ('here') attention path. Although conditional computation~\cite{conditional1,conditional2} and its application to attention mechanisms~\cite{colt5,moa} have been discussed in prior works, here we investigate the implications of enabling such dynamic sparsity within pre-trained decoder-only language models.

\subsection{Architecture}
\label{sec:architecture}
\begin{figure}
\centering
\includegraphics[width=0.6\textwidth]{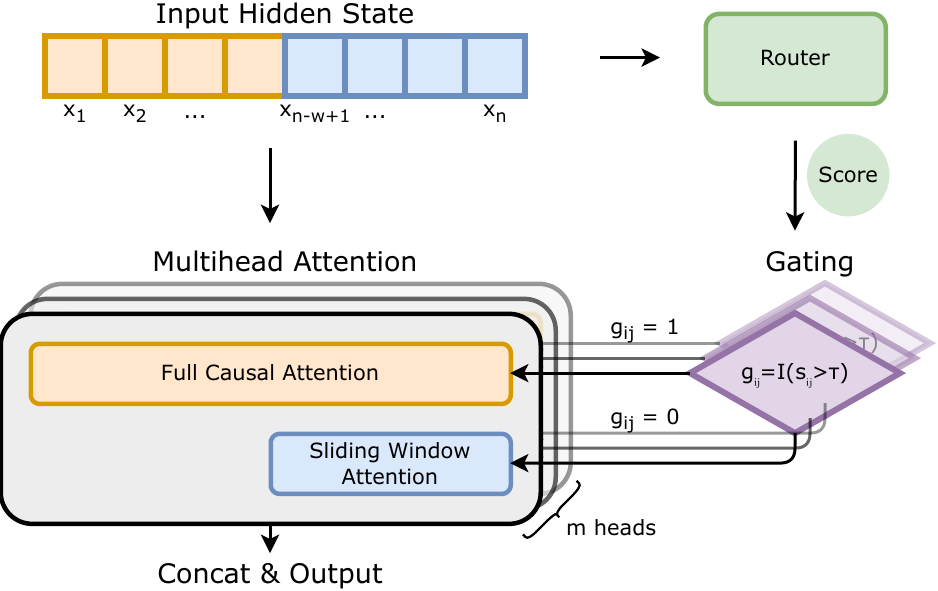}
\caption{Overview of the All-or-Here Attention (AHA) architecture. A lightweight router computes importance scores for each head, generating binary gates that dynamically toggle between full causal attention and local sliding window attention.}
\label{fig:fig2}
\end{figure}

Figure~\ref{fig:fig2} illustrates the pipeline within a single All-or-here Attention Block. The core structure is a dynamic routing mechanism that adaptively toggles between full attention and local sliding window attention.

Let the input hidden states be $\mathbf{X} \in \mathbb{R}^{n \times d}$, where $n$ is the sequence length and $d$ is the model dimension. The block comprises a total of $m$ attention heads, denoted by the set $\mathbf{H} = \{\text{head}_1, \text{head}_2, \dots, \text{head}_m\}$. To determine the attention scope for all tokens and all $m$ heads simultaneously, we employ a lightweight router represented by the weight matrix $\mathbf{W}_{\text{Router}} \in \mathbb{R}^{d \times m}$. The overall score matrix $\mathbf{S} \in \mathbb{R}^{n \times m}$ is computed by applying the sigmoid function to the linear projection of the input $\mathbf{X}$:
\begin{equation}
    \mathbf{S} = \sigma(\mathbf{X} \mathbf{W}_{\text{Router}}).
\label{eq:eq_router}
\end{equation}
The element $s_{i,j} \in (0, 1)$ of the matrix $\mathbf{S}$ represents the importance score for the $i$-th token within the $j$-th attention head. This score $s_{i,j}$ is then used to govern the binary decision for that specific token-head pair:
\begin{equation}
g_{i,j} = \mathbb{I}(s_{i,j} > \tau),
\label{eq:eq_gate}
\end{equation}
where $\mathbb{I}(\cdot)$ is the indicator function, and $\tau$ is a pre-defined threshold. The binary gate $g_{i,j}$ directly selects the attention mechanism: $g_{i,j} = 1$ triggers full attention, while $g_{i,j} = 0$ triggers local sliding window attention. 

To enable end-to-end optimization despite the non-differentiable nature of the discrete indicator function, we employ the Straight-Through Estimator (STE)~\cite{ste}. During the forward pass, the model utilizes the discrete decisions $g_{i,j}$ to strictly execute the selected attention path. During the backward pass, however, we approximate the thresholding operation as the identity function, allowing gradients to flow transparently from the gate to the score. Formally, the gradient of the loss $\mathcal{L}$ with respect to the router score $s_{i,j}$ is defined as:
\begin{equation}
    \frac{\partial \mathcal{L}}{\partial s_{i,j}} \approx \frac{\partial \mathcal{L}}{\partial g_{i,j}}.
\label{eq:ste}
\end{equation}
This allows gradients to directly update $\mathbf{W}_{\text{Router}}$, optimizing routing decisions for the global objective.

With the binary gate $g_{i,j}$ determined, the model conditionally executes the attention mechanism. Let $w$ denote the pre-defined size of the local sliding window. The attention output $\mathbf{a}_{i,j}$ for the $j$-th head at token step $i$ is formulated as:

\begin{equation}
    \mathbf{a}_{i,j} = 
    \begin{cases} 
    g_{i,j} \cdot \text{Attention}(\mathbf{q}_{i,j}, \mathbf{K}_{1:i, j}, \mathbf{V}_{1:i, j}) & \text{if } g_{i,j} = 1 \\
    (1-g_{i,j}) \cdot \text{Attention}(\mathbf{q}_{i,j}, \mathbf{K}_{i-w+1:i, j}, \mathbf{V}_{i-w+1:i, j}) & \text{if } g_{i,j} = 0 
    \end{cases}
\label{eq:attn_cases}
\end{equation}

where $\mathbf{q}_{i,j}$ is the query vector for the current token. The notation $\mathbf{K}_{1:i, j}$ represents the key states for the entire prefix (Full Attention), whereas $\mathbf{K}_{i-w+1:i, j}$ represents the truncated key states restricted to the most recent $w$ tokens (Sliding Window Attention). The function $\text{Attention}(\cdot)$ follows the standard scaled dot-product formulation. 

Following the conditional computation, we aggregate the results to form the final block output. For the current token $i$, the outputs from all $m$ heads are concatenated and projected via the output weight matrix $\mathbf{W}_O \in \mathbb{R}^{d \times d}$:
\begin{equation}
    \mathbf{O}_i = \text{Concat}(\mathbf{a}_{i,1}, \dots, \mathbf{a}_{i,m}) \mathbf{W}_O.
\label{eq:attn_output}
\end{equation}

Beyond the specific modifications to the attention mechanism, our architecture strictly adheres to the standard decoder-only Transformer design~\cite{gpt,llama}. Crucially, our method does not require pre-training from scratch. A standard, fully pre-trained Transformer can be seamlessly adapted by replacing its full attention modules with AHA, followed by continued Supervised Fine-Tuning (SFT). This capability allows AHA to serve as an drop-in enhancement for existing Large Language Models without incurring the prohibitive costs of full-scale pre-training.

\subsection{Loss Function}
\label{sec:loss}
During training, we optimize the model parameters using a joint objective function that balances standard language modeling performance with computational efficiency. The total loss $\mathcal{L}$ is defined as:
\begin{equation}
    \mathcal{L} = \mathcal{L}_{\text{LM}} + \lambda \mathcal{L}_{\text{reg}},
\label{eq:total_loss}
\end{equation}
where $\mathcal{L}_{\text{LM}}$ is the standard cross-entropy loss for next-token prediction, and $\lambda$ is a regularization hyperparameter controlling the trade-off between task accuracy and attention sparsity.

To encourage the model to prioritize the efficient local window attention, we apply an L1 penalty to the router scores across the entire network. Let $L$ denote the number of layers, $n$ the sequence length, and $m$ the number of attention heads. The regularization term is formulated as the mean of the router scores:
\begin{equation}
    \mathcal{L}_{\text{reg}} = \frac{1}{L \cdot n \cdot m} \sum_{k=1}^{L} \sum_{i=1}^{n} \sum_{j=1}^{m} s_{i,j}^{(k)},
\label{eq:reg_loss}
\end{equation}
where the superscript $(k)$ denotes the layer index. Minimizing this term drives the router scores $s_{i,j}^{(k)}$ towards zero. Since the gating decision is determined by $g_{i,j} = \mathbb{I}(s_{i,j} > \tau)$, reducing the scores effectively lowers the probability of triggering full attention. Consequently, this regularization incentivizes the model to default to the efficient sliding window mechanism, utilizing full attention only when necessary.

\section{Experiments}
\label{sec:experiments}

\subsection{Implementation Details}
To evaluate the efficacy of our proposed method, we implement the All-or-Here Attention (AHA) architecture using OLMo-2-0425-1B-SFT~\cite{olmo2} as the base model. We initialize the model with all original pre-trained parameters. In every attention block, we introduce the linear projection matrix $\mathbf{W}_{\text{Router}}$ to generate the routing score $s_{i,j}$ for the $i$-th token in the $j$-th head. The gating mechanism operates with a fixed threshold of $\tau=0.5$. regarding the training objective, we set the regularization coefficient to $\lambda=3 \times 10^{-4}$. We fine-tune the model for one epoch on the Tulu-v3 dataset~\cite{tuluv3}. Since the base model was originally fine-tuned on this same dataset, this experimental design effectively isolates the impact of AHA from potential data distribution shifts. Optimization is performed using AdamW with a learning rate of $3 \times 10^{-5}$, $\beta_1=0.9$, and $\beta_2=0.95$, alongside a linear warmup ratio of $0.03$ and a global batch size of $128$.

\subsection{Main Results}

We evaluate the proposed All-or-Here Attention (AHA) mechanism by training a suite of models with varying local sliding window sizes $w \in \{16, 32, 64, 128, 256\}$, while keeping all other implementation details and hyperparameters identical to the previous section. This setup allows us to investigate how the capacity of the local context influences the router's reliance on full attention. Crucially, we ensure that the window size $w$ used during inference strictly matches the configuration used during training.

\paragraph{Benchmarks.} We evaluate our method across a diverse set of tasks, categorized into single-token and multi-token generation. For single-token tasks, we include MMLU~\citep{mmlu} for massive multitask understanding, HellaSwag~\citep{hellaswag} for grounded reasoning, and CSQA~\citep{csqa} for commonsense question answering. For multi-token tasks, we evaluate on GSM8K~\citep{gsm8k} for multi-step mathematical reasoning, MBPP~\citep{mbpp} for Python code generation, and MultiNews~\citep{multinews} for multi-document summarization.

All evaluations are conducted using the \texttt{lm-evaluation-harness} toolkit~\citep{eval-harness}. We employ a 5-shot setting for MMLU, HellaSwag, CSQA, and GSM8K, and a zero-shot setting for MBPP and MultiNews. Metrics include accuracy (Acc) for MMLU and CSQA, normalized accuracy (Acc\_norm) for HellaSwag, exact-match (EM) for GSM8K, Pass@1 for MBPP, and ROUGE scores for MultiNews. To ensure the sliding window attention is strictly operative, we filter out samples where the input context length is shorter than the window size. The average input context lengths for these tasks are 755, 532, 332, 939, 675, and 2,196 tokens, respectively. For multi-token generation, the model generates an average of 95 tokens for GSM8K, 52 tokens for MBPP, and 157 tokens for MultiNews.

\begin{table}[t]
\centering
\footnotesize 
\setlength{\tabcolsep}{3pt} 
\renewcommand{\arraystretch}{0.9} 

\caption{Performance comparison of the proposed All-or-here Attention across various window sizes. "Vanilla" denotes the OLMo-2-0425-1B-SFT model using full attention.}
\label{tab:tab1}
\begin{tabular}{lccccccc} 
\toprule
\textbf{Method} & \textbf{MMLU} & \textbf{HellaSwag} & \textbf{CSQA} & \textbf{GSM8K} & \textbf{MBPP} & \textbf{News} & \textbf{Retain \%} \\
\midrule
Vanilla & 0.4056 & 0.6487 & 0.5307 & 0.3874 & 0.1560 & 0.2074 & 100.0\% \\
\midrule
$w=16$  & 0.3676 & 0.6400 & 0.4578 & 0.3980 & 0.1260 & 0.2014 & 92.7\% \\
$w=32$  & 0.3847 & 0.6410 & 0.4586 & 0.4102 & 0.1240 & 0.1814 & 92.2\% \\
$w=64$  & 0.3963 & 0.6431 & 0.4980 & 0.4121 & 0.1200 & 0.1983 & 94.9\% \\
$w=128$ & 0.4087 & 0.6445 & 0.4996 & 0.4291 & 0.1580 & 0.2053 & 100.9\% \\
$w=256$ & 0.4134 & 0.6520 & 0.5053 & 0.4632 & 0.1600 & 0.1975 & 102.5\% \\
\bottomrule
\end{tabular}
\end{table}

\begin{table}[t]
\centering
\footnotesize 
\setlength{\tabcolsep}{3pt} 
\renewcommand{\arraystretch}{0.9}
\caption{Average Full Attention usage of the proposed All-or-here Attention across various window sizes.}
\label{tab:tab2}
\begin{tabular}{lccccccc} 
\toprule
\textbf{Method} & \textbf{MMLU} & \textbf{HellaSwag} & \textbf{CSQA} & \textbf{GSM8K} & \textbf{MBPP} & \textbf{News} & \textbf{Avg \%} \\
\midrule
Vanilla & 100.0\% & 100.0\% & 100\% & 100\% & 100\% & 100\% & 100.0\% \\
\midrule
$w=16$  & 66.0\% & 71.8\% & 59.5\% & 41.5\% & 41.3\% & 36.0\% & 52.7\% \\
$w=32$  & 52.1\% & 58.0\% & 43.7\% & 35.2\% & 32.0\% & 27.1\% & 41.4\% \\
$w=64$  & 34.4\% & 38.3\% & 24.9\% & 24.0\% & 24.6\% & 22.5\% & 28.1\% \\
$w=128$ & 11.2\% & 11.8\% & 7.4\% & 11.3\% & 13.6\% & 14.4\% & 11.6\% \\
$w=256$ & 5.9\% & 6.9\% & 4.4\% & 5.7\% & 7.4\% & 10.0\% & 6.7\% \\
\bottomrule
\end{tabular}
\end{table}

Table~\ref{tab:tab1} summarizes the performance of the proposed All-or-here Attention across various window sizes. We observe a clear positive correlation between the window size and model performance. As the window size $w$ increases, the model's accuracy progressively recovers, eventually matching the vanilla result of OLMo-2-SFT. Specifically, with window sizes of $w=128$ and $w=256$, AHA maintains 100.9\% and 102.5\% of the Vanilla performance, respectively. Even under the extreme constraint of $w=16$, the model retains 92.7\% of the original performance.

Table~\ref{tab:tab2} details the sparsity patterns across different benchmarks. We define the average full attention usage $\mu_{f}$ as the mean of the binary gating values $g_{i,j}^{(k)}$ (Equation~\ref{eq:eq_gate}) averaged across all $L$ layers, $m$ heads, and $n$ generated tokens:
\begin{equation}
\mu_{f} = \frac{1}{L \cdot n \cdot m} \sum_{k=1}^{L} \sum_{i=1}^{n} \sum_{j=1}^{m} g_{i,j}^{(k)}.
\end{equation}
This metric directly reflects the frequency with which the model selects the full attention path. Analyzing the variations of $\mu_{f}$ presented in Table~\ref{tab:tab2}, the results underscore both the efficiency and the adaptability of our approach. First, in terms of efficiency, AHA demonstrates that the vast majority of full attention is redundant given sufficient local context. At $w=256$, the model successfully bypasses 93.3\% of the full attention operations. Our method also exhibits robust adaptability to the available window size. We observe a clear inverse correlation between window size and full attention usage. When the local window is severely restricted (e.g., $w=16$), the router effectively identifies the information deficit and compensates by increasing global access to 52.7\%. This dynamic adjustment allows the model to retain 92.7\% of the original quality even under extreme constraints, confirming that the gating mechanism is actively responsive to the token-level context needs rather than learning a static bias.

Finally, we analyze the relationship between the window size and the model's reliance on full attention based on the statistics in Table~\ref{tab:tab2}. We observe that the demand for full attention diminishes rapidly as the local window expands. A distinct turning point becomes evident between $w=128$ and $w=256$. As the window size reaches 128, the average full attention usage drops to 11.6\%, and further declines to a mere 6.7\% at $w=256$. This empirical evidence suggests that context dependencies exhibit a long-tailed nature: the vast majority of tokens depend primarily on the local area, while only a sparse minority necessitate attention to access the global context. We hypothesize that this mirrors the intrinsic structure of natural language, where most syntactic and anaphoric dependencies occur within short ranges, with only a sparse long‑range tail spanning hundreds of tokens~\citep{dependency1,dependency2,dependency3}.
\begin{figure}[t] \centering \includegraphics[width=0.7\textwidth]{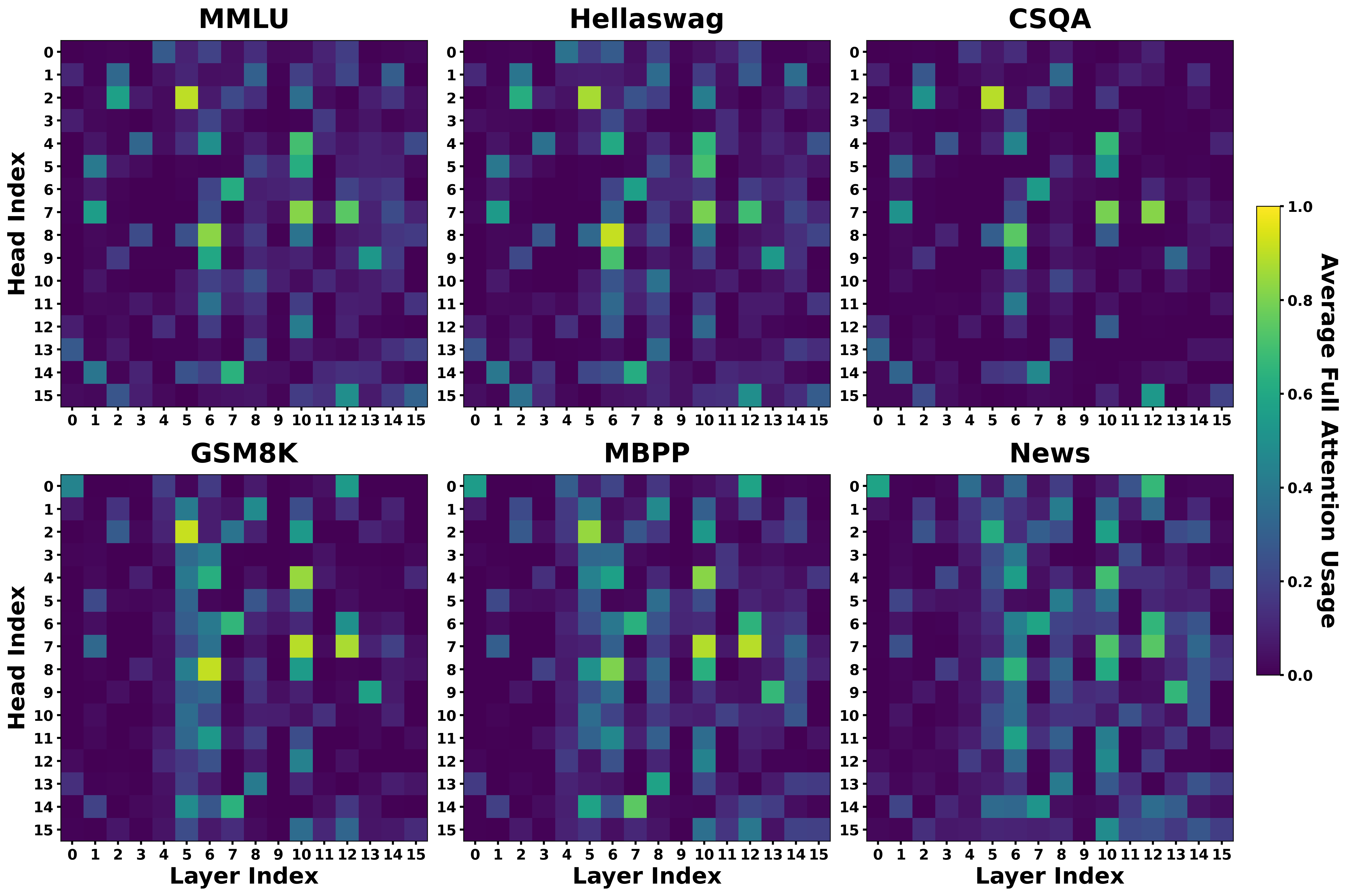} \caption{Visualization of the average full attention usage $\mu_f$ across different layers and heads. Lighter colors indicate a higher frequency of triggering full attention.} \label{fig:fig3} \end{figure}

\begin{figure}[t] \centering \includegraphics[width=0.8\textwidth]{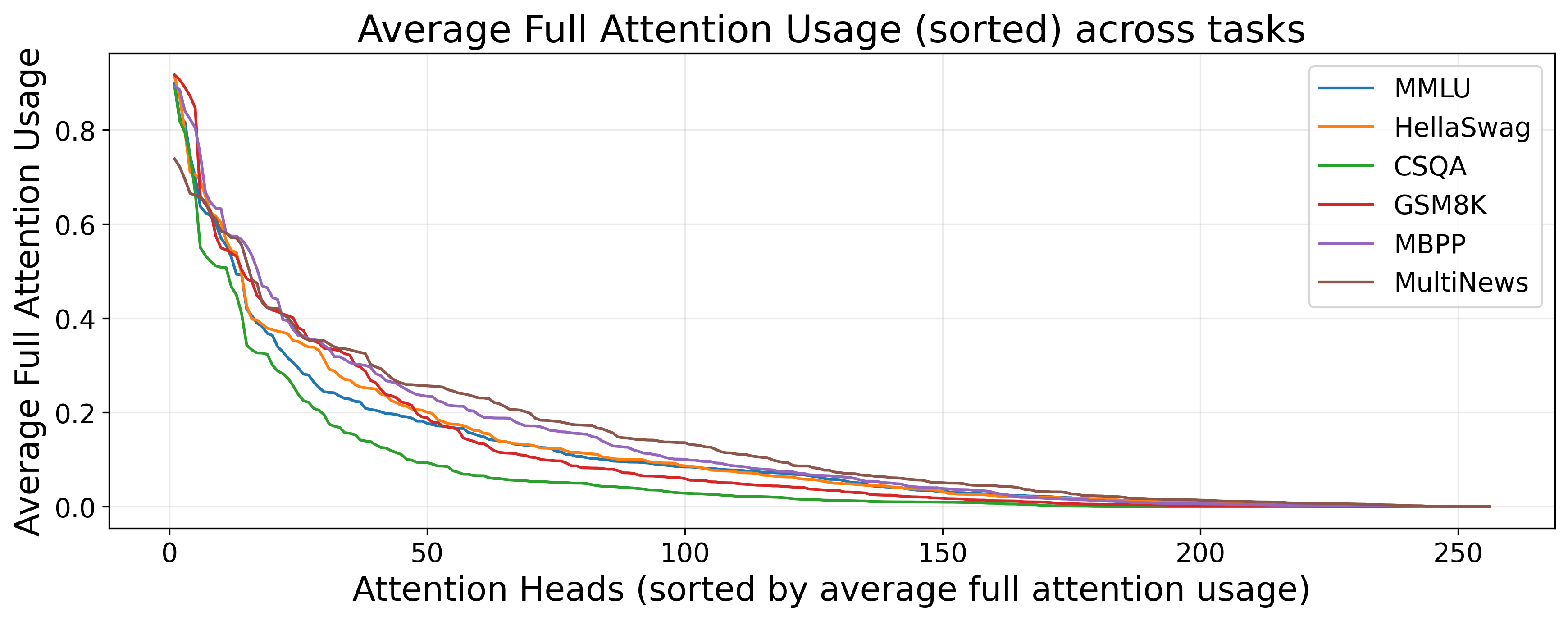} \caption{Sorted average full attention usage for all attention heads. The data shows that a small minority of heads account for the majority of full attention usage.} \label{fig:fig4} \end{figure}

\subsection{Sparsity Patterns and Head Specialization}

To investigate the internal mechanism of the proposed All-or-Here Attention, we visualize the average full attention usage $\mu_f$ for all attention heads across different layers and tasks in Figure~\ref{fig:fig3}. The heatmaps illustrate the frequency with which each head selects the full attention path. We observe two significant phenomena regarding the full attention usage.

First, the allocation of full attention exhibits a high degree of sparsity and follows a distinct long-tail distribution. As evident in the heatmaps, the majority of the grid remains dark, indicating that most attention heads rely almost exclusively on local context. To quantify this, we rank the attention heads by their average full attention usage $\mu_f$ in Figure~\ref{fig:fig4}. The results reveal a sharp decay: only a small fraction of "heavy-hitter" heads maintain high activation rates, while the curve drops precipitously, with the vast majority of heads effectively silencing their full attention path. This confirms that the model has learned a specialized division of labor, where broad global retrieval is concentrated in a few specific components, while the bulk of processing is handled locally.

Second, we identify the emergence of task-agnostic heads. Table 3 clarifies this by reporting the average interval, measured in tokens, between consecutive calls to full attention. We observe that certain heads, such as Layer 5, Head 2, maintain a gap consistently close to 1.0 across all benchmarks, signifying they attend globally for almost every token. Conversely, other components like Layer 3, Head 6 exhibit extreme sparsity, often processing thousands of tokens between full attention triggers. These results confirms that AHA effectively identifies a small subset of "always-on" global heads while allowing the majority of the model to operate within local constraints for thousands of tokens at a time.

\begin{table}[t]
\centering
\begin{tabular}{c|cccccc}
\toprule
\textbf{Head} & \textbf{MMLU} & \textbf{Hellaswag} & \textbf{CSQA} & \textbf{GSM8K } & \textbf{MBPP} & \textbf{NEWS}\\
\midrule
Layer 5, Head 2 & 1.11 & 1.15 & 1.12 & 1.09 & 1.07 & 1.04 \\
Layer 12, Head 8 & 8.83 & 22.23 & 525.43 & 56.07 & 27.68 &  2.02  \\
Layer 3, Head 6 & 427.54 & 1114.39 & 4154.94 & 10591.20 & 29228.29 & 268.54 \\
\bottomrule
\end{tabular}
\caption{Average full attention gap for representative heads across different benchmarks. The gap is defined as the mean number of tokens processed between consecutive triggers of the global attention path for a specific head.}
\label{tab:tab3}
\end{table}

\paragraph{Impact of Regularization Penalty.}To understand the trade-off between efficiency and performance, we train models with varying L1 regularization coefficients $\lambda \in \{1 \times 10^{-4}, 3 \times 10^{-4}, 1 \times 10^{-3}\}$ with a fixed window size of $w=128$. Table~\ref{tab:ablation_penalty} summarizes the results. We observe that the penalty value serves as a critical knob for balancing sparsity and capability. When the penalty is too aggressive ($\lambda=1 \times 10^{-3}$), the model is forced into extreme sparsity (only 4.5\% global attention). This leads to performance degradation, particularly on reasoning-intensive tasks such as GSM8K (dropping to 0.3730) and MBPP (dropping to 0.1320), indicating that complex reasoning requires a minimum threshold of global context. Conversely, relaxing the penalty ($\lambda=1 \times 10^{-4}$) drastically increases computational cost—utilizing 53.5\% global attention—without yielding meaningful performance gains. In fact, the accuracy on MMLU and News remains stagnant compared to the balanced setting ($\lambda=3 \times 10^{-4}$). This lack of improvement suggests a key hypothesis: a vast majority of full attention operations are fundamentally non-essential. Even when given the computational budget to attend globally, the model gains no additional signal, confirming that full attention is largely redundant and its excessive use may even introduce irrelevant noise. Our chosen default ($\lambda=3 \times 10^{-4}$) thus achieves a good Pareto frontier, maintaining high performance while discarding approximately 88\% of these redundant operations.

\begin{table}[t]
\centering
\footnotesize
\setlength{\tabcolsep}{3pt}
\renewcommand{\arraystretch}{0.9}
\caption{Ablation study on the L1 regularization penalty ($\lambda$). "Avg Full \%" denotes the mean average full attention usage across the tasks.}
\label{tab:ablation_penalty}
\begin{tabular}{lccccccc}
\toprule
\textbf{Penalty ($\lambda$)} & \textbf{MMLU} & \textbf{HellaSwag} & \textbf{CSQA} & \textbf{GSM8K} & \textbf{MBPP} & \textbf{News} & \textbf{Avg Full \%} \\
\midrule
$1 \times 10^{-3}$ & 0.4045 & 0.6419 & 0.4878 & 0.3730 & 0.1320 & 0.1997 & 4.5\% \\
$3 \times 10^{-4}$ & 0.4087 & 0.6445 & 0.4996 & 0.4291 & 0.1580 & 0.2053 & 11.6\% \\
$1 \times 10^{-4}$ & 0.4042 & 0.6456 & 0.4996 & 0.4397 & 0.1600 & 0.2047 & 53.5\% \\
\bottomrule
\end{tabular}
\end{table}

\section{Related Works}

\subsection{Efficient Attention}
To address the quadratic computational complexity of standard self-attention, efficient attention mechanisms are generally categorized into linear, sparse, and hierarchical approaches.  Linear attention reformulates the attention mechanism to scale linearly with sequence length, often employing kernel feature maps or recurrent state updates, as seen in the autoregressive framework of LinearAttn~\cite{linearattn} and the selective state-space models of Mamba~\cite{mamba}. Sparse attention reduces memory footprint by restricting token interactions to specific subsets. This includes static block-sparse patterns in Sparse Transformers~\cite{sparsetransformer} and structured sliding windows in Longformer~\cite{longformer} and BigBird~\cite{bigbird}. More advanced works have explored adaptive sparsity, ranging from Differentiable Attention Window~\cite{diffwindow} to Adaptive Attention Span~\cite{adaspan,fu2025mixture}, which dynamically determine the optimal context span for each head.  Finally, hierarchical attention organizes information processing across multiple levels of granularity, from the sentence-to-document aggregation in HAN~\cite{hierarchical} to the hardware-aligned, coarse-to-fine compression in Native Sparse Attention~\cite{deepseeknsa}.

In contrast to these increasingly sophisticated designs, our proposed All-or-Here Attention (AHA) demonstrates that complex span decision processes or hierarchical selection could be simplified when handling many tasks. We show that a minimalist binary routing decision---selecting strictly between local and global scopes---can eliminate over 90\% of full attention operations while maintaining model performance.

\subsection{Conditional Computation}

The notion of selectively skipping computation in deep neural networks can be traced back to conditional computation, which aims to reduce computational cost by activating only a subset of model components conditioned on the input~\cite{ste}. This idea has been extensively explored through mechanisms such as adaptive computation time~\cite{graves2017act}, mixture-of-experts~\cite{Shazeer2017MoE}, and mixture-of-depth~\cite{mod}. In Transformer-based architectures, prior work has shown that many attention heads are redundant or contribute unevenly across inputs, motivating approaches such as head pruning and head importance estimation ~\cite{Michel2019, voita-etal-2019-analyzing}. Complementary to these efforts, sparse and block-wise attention mechanisms restrict global attention to a subset of tokens or regions to improve efficiency~\cite{child2019, Beltagy2020Longformer, Zaheer2020}. More recent methods focus on how to make attention more efficient, e.g.,  FlashAttention~\cite{FlashAttention} and hierarchical attention~\cite{hierarchical,deepseeknsa}. Orthogonal to these studies, we demonstrate that in most cases global attention is not required at all, providing strong motivation for learning when and where to attend globally. 

\section{Limitations}
While our experiments empirically demonstrate that the vast majority of global attention operations are redundant, our analysis focuses on the resulting attention sparsity rather than system-level metrics such as wall-clock speedup. Realizing the practical acceleration from this sparsity presents non-trivial engineering challenges. Current hardware accelerators and standard attention kernels (e.g., FlashAttention~\cite{flashattn}) are heavily optimized for static, dense matrix operations. In contrast, AHA introduces dynamic, heterogeneous execution paths where different heads process varying context lengths within the same batch. Consequently, we position AHA as a proof-of-concept for algorithmic redundancy, leaving the development of hardware-aware kernels to leverage this sparsity for future work. Furthermore, our current approach employs a strictly binary routing mechanism. It is worth extending this design to support a multi-choice decision space, enabling the model to dynamically select among global attention and multiple local windows of varying sizes, potentially integrated with paradigms like Mixture of Attention Spans~\cite{fu2025mixture}.

Due to limitations in both lab computational resources and available training data, we conducted experiments only on OLMo-2, which uses full attention. It would be interesting to extend our evaluation to other attention mechanisms. Since index-based attention,  top-k attention, sparse attention, and hierarchical attention are all approximations of full attention, if full attention itself does not need to be activated frequently, it is reasonable to expect that these approximations would likewise remain inactive even if included. However, this hypothesis requires further experimental verification.

\section{Conclusion}

In this paper, we introduced All-or-Here Attention (AHA), a simple yet effective paradigm that dynamically toggles between efficient local sliding window attention and full global attention at the token level. Unlike previous methods that rely on soft gating or complex approximations, our approach explicitly decouples local processing from global retrieval through strict conditional computation. Our empirical results on OLMo-2 confirm that the vast majority of global attention operations are redundant. We demonstrate that with a window size of 256 tokens, AHA can eliminate over 95\% of full attention calculations without compromising performance on standard benchmarks. Furthermore, our analysis reveals that context dependency follows a long-tail distribution: as the local receptive field expands, the need for global access decays rapidly. These findings indicate that the key to efficiency lies not in compressing the full attention matrix, but in accurately identifying the sparse moments when global context is truly required. We hope our findings encourage further exploration of simple, dynamic strategies for efficient language modeling.

\section*{Acknowledgements}
We would like to gratefully acknowledge the generous support of the NVIDIA Academic Grant Program. Access to NVIDIA GPUs and software toolkits enabled us to conduct experiments on large training datasets and inspired new research directions.

\newpage
\bibliography{references}

@article{attention,
  title={Neural Machine Translation by Jointly Learning to Align and Translate},
  author={Dzmitry Bahdanau and Kyunghyun Cho and Yoshua Bengio},
  journal={CoRR},
  year={2014},
}

@inproceedings{transformer,
  title={Attention is All you Need},
  author={Ashish Vaswani and Noam M. Shazeer and Niki Parmar and Jakob Uszkoreit and Llion Jones and Aidan N. Gomez and Lukasz Kaiser and Illia Polosukhin},
  booktitle={Neural Information Processing Systems},
  year={2017},
}

@inproceedings{bert,
  title={BERT: Pre-training of Deep Bidirectional Transformers for Language Understanding},
  author={Jacob Devlin and Ming-Wei Chang and Kenton Lee and Kristina Toutanova},
  booktitle={North American Chapter of the Association for Computational Linguistics},
  year={2019},
}

@article{gpt,
  title={Language Models are Few-Shot Learners},
  author={Tom B. Brown and Benjamin Mann and Nick Ryder and others},
  journal={ArXiv},
  year={2020},
}

@article{instructgpt,
  title={Training language models to follow instructions with human feedback},
  author={Long Ouyang and Jeff Wu and Xu Jiang and others},
  journal={ArXiv},
  year={2022},
}

@article{llama,
  title={LLaMA: Open and Efficient Foundation Language Models},
  author={Hugo Touvron and Thibaut Lavril and Gautier Izacard and others},
  journal={ArXiv},
  year={2023},
}

@article{qwen3,
  title={Qwen3 Technical Report},
  author={An Yang and Anfeng Li and Baosong Yang and others},
  journal={ArXiv},
  year={2025},
}

@inproceedings{linearattn,
  title={Transformers are RNNs: Fast Autoregressive Transformers with Linear Attention},
  author={Angelos Katharopoulos and Apoorv Vyas and Nikolaos Pappas and Franccois Fleuret},
  booktitle={International Conference on Machine Learning},
  year={2020},
}

@inproceedings{
mamba,
title={Mamba: Linear-Time Sequence Modeling with Selective State Spaces},
author={Albert Gu and Tri Dao},
booktitle={First Conference on Language Modeling},
year={2024},
}

@article{gateddeltanet,
  title={Gated Delta Networks: Improving Mamba2 with Delta Rule},
  author={Songlin Yang and Jan Kautz and Ali Hatamizadeh},
  journal={ArXiv},
  year={2024},
}

@inproceedings{
gatedattention,
title={Gated Attention for Large Language Models: Non-linearity, Sparsity, and Attention-Sink-Free},
author={Zihan Qiu and Zekun Wang and Bo Zheng and others},
booktitle={The Thirty-ninth Annual Conference on Neural Information Processing Systems},
year={2025},
}

@article{bigbird,
  title={Big Bird: Transformers for Longer Sequences},
  author={Manzil Zaheer and Guru Guruganesh and Kumar Avinava Dubey and others},
  journal={ArXiv},
  year={2020},
}

@article{longformer,
  title={Longformer: The Long-Document Transformer},
  author={Iz Beltagy and Matthew E. Peters and Arman Cohan},
  journal={ArXiv},
  year={2020},
}

@article{sparsetransformer,
  title={Generating Long Sequences with Sparse Transformers},
  author={Rewon Child and Scott Gray and Alec Radford and Ilya Sutskever},
  journal={ArXiv},
  year={2019},
}

@inproceedings{hierarchical,
  title={Hierarchical Attention Networks for Document Classification},
  author={Zichao Yang and Diyi Yang and Chris Dyer and Xiaodong He and Alex Smola and Eduard H. Hovy},
  booktitle={North American Chapter of the Association for Computational Linguistics},
  year={2016},
}

@inproceedings{deepseeknsa,
  title={Native Sparse Attention: Hardware-Aligned and Natively Trainable Sparse Attention},
  author={Jingyang Yuan and Huazuo Gao and Damai Dai and others},
  booktitle={Annual Meeting of the Association for Computational Linguistics},
  year={2025},
}

@article{ste,
  title={Estimating or Propagating Gradients Through Stochastic Neurons for Conditional Computation},
  author={Yoshua Bengio and Nicholas L{\'e}onard and Aaron C. Courville},
  journal={ArXiv},
  year={2013},
}

@article{dependency1,
  title={Dependency distance: A new perspective on syntactic patterns in natural languages},
  author={Liu, Haitao and Xu, Chunshan and Liang, Junying},
  journal={Physics of life reviews},
  year={2017},
}

@article{dependency2,
  title={Probability distribution of dependency distance and dependency type in translational language},
  author={Fan, Lu and Jiang, Yue},
  journal={Humanities and Social Sciences Communications},
  year={2023},
}

@article{dependency3,
  title={Large-scale evidence of dependency length minimization in 37 languages},
  author={Richard Futrell and Kyle Mahowald and Edward Gibson},
  journal={Proceedings of the National Academy of Sciences},
  year={2015},
}

@article{olmo2,
  title={2 OLMo 2 Furious},
  author={Team OLMo and Pete Walsh and Luca Soldaini and Dirk Groeneveld and others},
  journal={ArXiv},
  year={2024},
}

@article{tuluv3,
  title={T{\"U}LU 3: Pushing Frontiers in Open Language Model Post-Training},
  author={Nathan Lambert and Jacob Daniel Morrison and Valentina Pyatkin and others},
  journal={ArXiv},
  year={2024},
}

@misc{eval-harness,
  author       = {Gao, Leo, Tow and others},
  title        = {A framework for few-shot language model evaluation},
  year         = {2024},
  version      = {v0.5.1},
}

@inproceedings{mmlu,
  author       = {Dan Hendrycks and Collin Burns and Steven Basart and Andy Zou and Mantas Mazeika and Dawn Song and Jacob Steinhardt},
  title        = {Measuring Massive Multitask Language Understanding},
  booktitle    = {9th International Conference on Learning Representations, {ICLR} 2021, Virtual Event, Austria, May 3-7, 2021},
  year         = {2021}
}

@inproceedings{hellaswag,
  author       = {Rowan Zellers and Ari Holtzman and Yonatan Bisk and Ali Farhadi and Yejin Choi},
  title        = {HellaSwag: Can a Machine Really Finish Your Sentence?},
  booktitle    = {Proceedings of the 57th Conference of the Association for Computational Linguistics, {ACL} 2019, Florence, Italy, July 28- August 2, 2019, Volume 1: Long Papers},
  year         = {2019}
}

@article{gsm8k,
  title={Training Verifiers to Solve Math Word Problems},
  author={Karl Cobbe and Vineet Kosaraju and Mo Bavarian and others},
  journal={ArXiv},
  year={2021},
}

@article{mbpp,
  title={Program Synthesis with Large Language Models},
  author={Jacob Austin and Augustus Odena and Maxwell Nye and others},
  journal={ArXiv},
  year={2021},
}

@article{csqa,
  title={CommonsenseQA: A Question Answering Challenge Targeting Commonsense Knowledge},
  author={Alon Talmor and Jonathan Herzig and Nicholas Lourie and Jonathan Berant},
  journal={ArXiv},
  year={2019},
}

@inproceedings{multinews,
  title={Multi-News: A Large-Scale Multi-Document Summarization Dataset and Abstractive Hierarchical Model},
  author={Alexander R. Fabbri and Irene Li and Tianwei She and others},
  booktitle={Annual Meeting of the Association for Computational Linguistics},
  year={2019},
  url={https://api.semanticscholar.org/CorpusID:174799390}
}

@article{diffwindow,
  title={Differentiable Window for Dynamic Local Attention},
  author={Thanh-Tung Nguyen and Xuan-Phi Nguyen and Shafiq R. Joty and Xiaoli Li},
  journal={ArXiv},
  year={2020},
}

@article{adaspan,
  title={adaspan},
  author={Sainbayar Sukhbaatar and Edouard Grave and Piotr Bojanowski and Armand Joulin},
  journal={ArXiv},
  year={2019},
}

@misc{deepseekv32,
      title={DeepSeek-V3.2: Pushing the Frontier of Open Large Language Models}, 
      author={DeepSeek-AI},
      journal={ArXiv},
      year={2025},
}

@inproceedings{colt5,
  author       = {Joshua Ainslie and
                  Tao Lei and
                  Michiel de Jong and
                  others},
  title        = {CoLT5: Faster Long-Range Transformers with Conditional Computation},
  booktitle    = {Proceedings of the 2023 Conference on Empirical Methods in Natural
                  Language Processing, {EMNLP} 2023, Singapore, December 6-10, 2023},
  year         = {2023},
}

@article{conditional1,
  author       = {Emmanuel Bengio and
                  Pierre{-}Luc Bacon and
                  Joelle Pineau and
                  Doina Precup},
  title        = {Conditional Computation in Neural Networks for faster models},
  journal      = {CoRR},
  year         = {2015},
}

@article{conditional2,
  author       = {Ankur Bapna and
                  Naveen Arivazhagan and
                  Orhan Firat},
  title        = {Controlling Computation versus Quality for Neural Sequence Models},
  journal      = {CoRR},
  year         = {2020},
}

@inproceedings{moa,
  author       = {Xiaofeng Zhang and
                  Yikang Shen and
                  Zeyu Huang and
                  others},
  title        = {Mixture of Attention Heads: Selecting Attention Heads Per Token},
  booktitle    = {Proceedings of the 2022 Conference on Empirical Methods in Natural
                  Language Processing, {EMNLP} 2022, Abu Dhabi, United Arab Emirates,
                  December 7-11, 2022},
  year         = {2022},
}

@misc{graves2017act,
      title={Adaptive Computation Time for Recurrent Neural Networks}, 
      author={Alex Graves},
      year={2017},
      eprint={1603.08983},
      archivePrefix={arXiv},
      primaryClass={cs.NE} 
}

@misc{Shazeer2017MoE,
      title={Outrageously Large Neural Networks: The Sparsely-Gated Mixture-of-Experts Layer}, 
      author={Noam Shazeer and Azalia Mirhoseini and Krzysztof Maziarz and Andy Davis and Quoc Le and Geoffrey Hinton and Jeff Dean},
      year={2017},
      eprint={1701.06538},
      archivePrefix={arXiv},
      primaryClass={cs.LG}
}

@inproceedings{Michel2019,
 author = {Michel, Paul and Levy, Omer and Neubig, Graham},
 booktitle = {Advances in Neural Information Processing Systems},
 editor = {H. Wallach and H. Larochelle and A. Beygelzimer and F. d\textquotesingle Alch\'{e}-Buc and E. Fox and R. Garnett},
 title = {Are Sixteen Heads Really Better than One?},
 volume = {32},
 year = {2019}
}

@inproceedings{voita-etal-2019-analyzing,
    title = "Analyzing Multi-Head Self-Attention: Specialized Heads Do the Heavy Lifting, the Rest Can Be Pruned",
    author = "Voita, Elena  and
      Talbot, David  and
      Moiseev, Fedor  and
      Sennrich, Rico  and
      Titov, Ivan",
    booktitle = "Proceedings of the 57th Annual Meeting of the Association for Computational Linguistics",
    month = jul,
    year = "2019",
    pages = "5797--5808",
   
}

@misc{child2019,
      title={Generating Long Sequences with Sparse Transformers}, 
      author={Rewon Child and Scott Gray and Alec Radford and Ilya Sutskever},
      year={2019},
      eprint={1904.10509},
      archivePrefix={arXiv},
      primaryClass={cs.LG}
}

@article{Beltagy2020Longformer,
  title={Longformer: The Long-Document Transformer},
  author={Iz Beltagy and Matthew E. Peters and Arman Cohan},
  journal={arXiv:2004.05150},
  year={2020},
}

@inproceedings{Zaheer2020,
author = {Zaheer, Manzil and Guruganesh, Guru and Dubey, Avinava and Ainslie, Joshua and Alberti, Chris and Ontanon, Santiago and Pham, Philip and Ravula, Anirudh and Wang, Qifan and Yang, Li and Ahmed, Amr},
title = {Big bird: transformers for longer sequences},
year = {2020},
isbn = {9781713829546},
publisher = {Curran Associates Inc.},
address = {Red Hook, NY, USA},
booktitle = {Proceedings of the 34th International Conference on Neural Information Processing Systems},
articleno = {1450},
numpages = {15},
series = {NIPS '20}
}

@inproceedings{FlashAttention,
author = {Dao, Tri and Fu, Daniel Y. and Ermon, Stefano and Rudra, Atri and R\'{e}, Christopher},
title = {FLASHATTENTION: fast and memory-efficient exact attention with IO-awareness},
year = {2022},
booktitle = {Proceedings of the 36th International Conference on Neural Information Processing Systems},
articleno = {1189},
numpages = {16},
series = {NIPS '22}
}

@inproceedings{
fu2025mixture,
title={Mixture of Attention Spans: Optimizing {LLM} Inference Efficiency with Heterogeneous Sliding-Window Lengths},
author={Tianyu Fu and Haofeng Huang and Xuefei Ning and Genghan Zhang and Boju Chen and Tianqi Wu and Hongyi Wang and Zixiao Huang and Shiyao Li and Shengen Yan and Guohao Dai and Huazhong Yang and Yu Wang},
booktitle={Second Conference on Language Modeling},
year={2025}
}

@inproceedings{flashattn,
  author       = {Tri Dao and
                  Daniel Y. Fu and
                  Stefano Ermon and
                  Atri Rudra and
                  Christopher R{\'{e}}},
  title        = {FlashAttention: Fast and Memory-Efficient Exact Attention with IO-Awareness},
  booktitle    = {Advances in Neural Information Processing Systems 35: Annual Conference
                  on Neural Information Processing Systems 2022, NeurIPS 2022, New Orleans,
                  LA, USA, November 28 - December 9, 2022},
  year         = {2022},
}

@article{mod,
  author    = {David Raposo and Samuel Ritter and Blake A. Richards and Timothy P. Lillicrap and Peter Conway Humphreys and Adam Santoro},
  title     = {Mixture-of-Depths: Dynamically allocating compute in transformer-based language models},
  journal   = {CoRR},
  year      = {2024},
  eprinttype= {arXiv}
}
\bibliographystyle{plain}
\end{document}